\pdfoutput=1

\documentclass[11pt]{article}

\usepackage[preprint]{acl}

\usepackage{times}
\usepackage{latexsym}

\usepackage[T1]{fontenc}

\usepackage[utf8]{inputenc}

\usepackage{microtype}

\usepackage{inconsolata}

\usepackage{graphicx}
\usepackage{multirow}

\title{Increasing the Robustness of the Fine-tuned Multilingual Machine-Generated Text Detectors}

\author{Dominik Macko, Robert Moro, Ivan Srba \\
  Kempelen Institute of Intelligent Technologies, Slovakia\\
  \texttt{\{dominik.macko, robert.moro, ivan.srba\}}@kinit.sk \\}

\begin{document}
\maketitle
\begin{abstract}
Since the proliferation of LLMs, there have been concerns about their misuse for harmful content creation and spreading. Recent studies justify such fears, providing evidence of LLM vulnerabilities and high potential of their misuse. Humans are no longer able to distinguish between high-quality machine-generated and authentic human-written texts. Therefore, it is crucial to develop automated means to accurately detect machine-generated content. It would enable to identify such content in online information space, thus providing an additional information about its credibility. This work addresses the problem by proposing a robust fine-tuning process of LLMs for the detection task, making the detectors more robust against obfuscation and more generalizable to out-of-distribution data.
\end{abstract}

\section{Introduction}

The number, variety, enhanced capabilities, and simple usage of large language models (LLMs) cause growing concerns about a massive generation of harmful content with LLMs being potentially maliciously misused ~\citep{borji2023categoricalarchivechatgptfailures, zhuo2023redteamingchatgptjailbreaking, crothers2023machine}. Previous research have demonstrated the ability of LLMs to produce disinformation texts~\citep{vykopal-etal-2024-disinformation, lucas-etal-2023-fighting}, even personalized ones~\cite{zugecova2024evaluationllmvulnerabilitiesmisused, gabriel2024generative}, frauds for academic exams \citep{openai2023gpt4}, or plagiarism \citep{wahle-etal-2022-large}. The fears of misuse are more severe due to indistinguishability between human-written texts and high-quality machine-generated texts \citep{sadasivan2025can}. Thus, the machine-generated text detection (MGTD) belongs to the key challenges associated with state-of-the-art LLMs as identified by \citep{kaddour2023challenges}.

Since the harmful online content, such as misinformation, disinformation, propaganda, or impersonation, can be also generated or adjusted by LLMs, its automated detection can crucially help to fight such phenomena. Indication that a potentially harmful content is generated by using artificial intelligence (AI) can alleviate its eventual effect on people. By its statistical nature, the AI generated content should be considered less credible (in comparison to authentic human-written texts), mainly due to the simplicity and speed of creating the content as well as a potential presence of hallucinations (unintended misleading information).

However, the state-of-the-art MGTD methods suffer by performance degradation in out-of-distribution settings, such as new generators, domains, languages, and paraphrasing or adversarial attacks \citep{antoun-etal-2023-towards, li-etal-2024-mage}. Therefore, increasing their robustness in this manner is the primary concern.

This work addresses this problem by mixing the training data in a way to include multiple languages, domains, generators and even obfuscated (attack) data. To the best of our knowledge, this is the first attempt to systematically fine-tune robust multilingual MGT detectors (the source code available\footnote{\scriptsize\url{https://anonymous.4open.science/r/robust_mgt_detector}}). Although such detectors manifest a small performance degradation on in-distribution data, the out-of-distribution performance is significantly increased, making them more suitable to be deployed in the real-world settings.

\section{Datasets}

For the detectors training, we have used a combination of multiple datasets described below. The statistics regarding sample counts for training are summarized in Table~\ref{tab:datasets}.

\begin{table}[!t]
\centering
\resizebox{\linewidth}{!}{
\begin{tabular}{p{3.1cm}|cccc}
\hline
\bfseries Dataset & \bfseries Human & \bfseries Machine & \bfseries Generator & \bfseries Language \\
\hline
\textbf{MultiSocial} & 290,618 & 290,618 & 7 & 40 \\
\textbf{MULTITuDE\_v2} & 0 & 120,090 & 7 & 3 \\
\textbf{MULTITuDE\_v3} & 136,577 & 143,527 & 7 & 22 \\
\textbf{MIX2k} & 1,000 & 1,000 & 75 & 7 \\
\hline
\end{tabular}
}
\vspace{-2mm}
\caption{Text samples counts for the train splits of the included datasets (MIX2k used for validation only).}
\label{tab:datasets}
\vspace{-5mm}
\end{table}

\paragraph{MultiSocial} \citep{multisocial}
A dataset of real social-media texts in 22 languages from 5 platforms, containing also the generated counterparts (by using 3 iterations of paraphrasing) by 7 generators. We have balanced the train split of this dataset by additional human samples (using the published procedure\footnote{\scriptsize\url{https://anonymous.4open.science/r/multisocial}}), covering also languages out of the intended 22 (specifically, 40 languages with at least 1,000 samples).

\paragraph{MULTITuDE\_v2} \citep{macko-etal-2024-authorship}
A subset of MULTITuDE\_v2 dataset of obfuscated MULTITuDE \citep{macko-etal-2023-multitude} texts, covering the three selected authorship obfuscation methods (DFTFooler adversarial attack, ChatGPT paraphrasing, and m2m100-1.2B backtranslation). Due to modification (obfuscation) of the texts by machines, we include only machine-labeled samples.

\paragraph{MULTITuDE\_v3} \citep{multisocial}
An extended version of the MULTITuDE \citep{macko-etal-2023-multitude} dataset of real human-written news articles in 22 languages accompanied by counterparts generated by 7 generators, created for out-of-distribution evaluation by \citep{multisocial}. We have requested this dataset from the authors (since it is not published) and we have balanced their train split of the dataset by additional human samples (using the same published procedure\footnote{\scriptsize\url{https://github.com/kinit-sk/mgt-detection-benchmark}}).

\paragraph{MIX2k}
Our 2k subset of a mixture of existing datasets on binary machine-generated text detection task, covering various domains, 7 languages and over 75 generators (for more information about the dataset creation process, see Appendix~\ref{sec:appendix_mix}). It contains 1,000 human samples and 1,000 machine samples.

We have combined the train splits of the first three datasets for detectors fine-tuning. After de-duplication, it includes 427,195 human samples and 554,216 machine samples out of 2 domains (news and social media), 44 languages (of more than 1,000 samples), and 16 generators. A small imbalance in human vs machine classes was dealt by up-sampling of the minority class (i.e., human) during the training. The MIX2k dataset (containing mostly out-of-distribution data) was used for validation during the fine-tuning, i.e., for checkpoint selection.

\section{Fine-tuning Procedure}

We used two kinds of fine-tuning, based on the model size. For smaller models, we have used the full fine-tuning process utilizing half precision, with the default AdamW optimizer with learning rate of $2e-6$, batch size of 32, validation each 2,000 steps, running at maximum for 10 epochs or 4 days.

For the models having more than 2B parameters, we have used QLoRA \citep{dettmers2024qlora} parameter efficient fine-tuning process with the weighted cross entropy with logits for loss calculations and 4-bit quantization using BitsAndBytes\footnote{\scriptsize\url{https://pypi.org/project/bitsandbytes}}. The LoRA configuration\footnote{\scriptsize\url{https://pypi.org/project/peft}} used an \textit{alpha} of 16, a \textit{dropout} of 0.1, \textit{r} of 64, and the \textit{task type} of sequence classification targeting all linear modules. In comparison to full fine-tuning, the QLoRA fine-tuning version used paged AdamW optimizer with learning rate of $2e-5$, gradient accumulation of 4 steps, and evaluation each 200 steps.

We used these hyperparameter values for all models to make them comparable. However, if a smaller batch size was necessary due to memory constraints in case of bigger models, we tuned the values accordingly.

\section{Experimental Settings}

As a baseline, we have used the mDeBERTa-v3-base model (as shown by existing works as the best of smaller pre-trained models) fine-tuned on a combination of MULTITuDE\_v3 and MultiSocial datasets (train splits without our balanced extension by additional human samples).

We have selected 5 models with diverse architectures and numbers of parameters for robust fine-tuning. 

For the in-distribution evaluation, we have used the test splits of the MULTITuDE\_v3 and MultiSocial datasets. For the out-of-distribution evaluation, we have used the test split of the SemEval-2024 Task 8 multilingual dataset \citep{wang-etal-2024-semeval-2024}, as well as the full MIX dataset (200k samples). The statistics regarding sample counts for testing are summarized in Table~\ref{tab:datasets_test}.

\begin{table}[!t]
\centering
\resizebox{\linewidth}{!}{
\begin{tabular}{p{3.1cm}|cccc}
\hline
\bfseries Dataset & \bfseries Human & \bfseries Machine & \bfseries Generator & \bfseries Language \\
\hline
\textbf{MultiSocial} & 17,367 & 121,460 & 7 & 22 \\
\textbf{MULTITuDE\_v3} & 6,502 & 44,566 & 7 & 22 \\
\textbf{MIX} & 99,759 & 100,000 & 75 & 7 \\
\textbf{SemEval} & 20,238 & 22,140 & 7 & 4 \\
\hline
\end{tabular}
}
\vspace{-2mm}
\caption{Text samples counts for the test splits of the evaluation datasets.}
\label{tab:datasets_test}
\vspace{-5mm}
\end{table}

As the evaluation metric, we are primarily using \textbf{AUC ROC} (area under curve of the receiver operating characteristic) as a classification-threshold independent metric reflecting generic detection capability. For a fine-grained comparison, we are also using \textbf{TPR @ 5\% FPR}, representing true positive rate (\textbf{TPR}) using the classification threshold calibrated to reach 5\% of false positive rate (\textbf{FPR}) based on ROC curve, and \textbf{Macro F1} (macro average of the F1 score) reflecting a balance between precision and recall metrics.

\section{Results}

When evaluating on in-distribution data (Table~\ref{tab:indistribution}), we can notice the highest performance of the baseline detector in both AUC ROC and TPR @ 5\% FPR metrics in news as well as social-media texts. This makes the baseline really strong. However, we hypothesize that such a baseline, trained purely on the train splits of the evaluation datasets, suffers by over-fitting to the data distribution and thus generalizes worse to out-of-distribution data.

\begin{table}[!b]
\vspace{-3mm}
\centering
\resizebox{\linewidth}{!}{
\begin{tabular}{l|cc|cc}
\hline
 & \multicolumn{2}{c|}{\textbf{MULTITuDE\_v3}} & \multicolumn{2}{c}{\textbf{MultiSocial}} \\
\textbf{Detector} & \textbf{AUC ROC} & \textbf{TPR @ 5\% FPR} & \textbf{AUC ROC} & \textbf{TPR @ 5\% FPR} \\
\hline
Gemma-2-9b-it & 0.9914 & 0.9798 & 0.9563 & 0.8400 \\
Qwen2-0.5B & 0.9785 & 0.9316 & 0.9582 & 0.8413 \\
Qwen2-1.5B & 0.9883 & 0.9460 & 0.9549 & 0.7781 \\
Yi-1.5-6B & 0.9748 & 0.9115 & 0.9474 & 0.8008 \\
mDeBERTa-v3-base & \textbf{0.9959} & 0.9797 & 0.9540 & 0.7750 \\
mDeBERTa-v3-base (baseline) & 0.9944 & \textbf{0.9875} & \textbf{0.9746} & \textbf{0.8862} \\
\hline
\end{tabular}
}
\vspace{-2mm}
\caption{Evaluation of detection performance on in-distribution data. The highest value in each column is boldfaced.}
\label{tab:indistribution}
\end{table}
\begin{table}[!b]
\vspace{-3mm}
\centering
\resizebox{\linewidth}{!}{
\begin{tabular}{l|cc|cc}
\hline
 & \multicolumn{2}{c|}{\textbf{MIX}} & \multicolumn{2}{c}{\textbf{SemEval}} \\
\textbf{Detector} & \textbf{AUC ROC} & \textbf{TPR @ 5\% FPR} & \textbf{AUC ROC} & \textbf{TPR @ 5\% FPR} \\
\hline
Gemma-2-9b-it & \bfseries 0.8901 & \bfseries 0.5227 & \bfseries 0.9448 & 0.8284 \\
Qwen2-0.5B & 0.6499 & 0.0000 & 0.8434 & 0.0000 \\
Qwen2-1.5B & 0.7588 & 0.3064 & 0.9391 & \bfseries 0.8287 \\
Yi-1.5-6B & 0.8167 & 0.0000 & 0.8922 & 0.0000 \\
mDeBERTa-v3-base & 0.6669 & 0.0502 & 0.8666 & 0.6843 \\
mDeBERTa-v3-base (baseline) & 0.5502 & 0.0000 & 0.8305 & 0.0000 \\
\hline
\end{tabular}
}
\vspace{-2mm}
\caption{Evaluation of detection performance on out-of-distribution data. The highest value in each column is boldfaced.}
\label{tab:outofdistribution}
\end{table}

All of the detectors are performing better on the MULTITuDE\_v3 news articles than on the Multisocial social-media texts, where the lowest performance is achieved by robustly fine-tuned mDeBERTa-v3-base and Qwen2-1.5B models.

When looking at the per-language AUC ROC performance (Table~\ref{tab:indistribution_lang}), we can see that the detection works well for all the languages evenly, with a small exception of surprise languages (ga, gd, sl) in MultiSocial (i.e., not enough social-media texts for training in some languages). This further evokes a lower generalizability to out-of-distribution data.

\begin{table*}[!t]
\centering
\resizebox{\textwidth}{!}{
\addtolength{\tabcolsep}{-2pt}
\begin{tabular}{cp{5cm}|cccccccccccccccccccccc}
\hline
& & \multicolumn{22}{c}{\bfseries Test Language [AUC ROC]} \\
\textbf{Dataset} & \bfseries Detector & \bfseries ar & \bfseries bg & \bfseries ca & \bfseries cs & \bfseries de & \bfseries el & \bfseries en & \bfseries es & \bfseries et & \bfseries ga & \bfseries gd & \bfseries hr & \bfseries hu & \bfseries nl & \bfseries pl & \bfseries pt & \bfseries ro & \bfseries ru & \bfseries sk & \bfseries sl & \bfseries uk & \bfseries zh \\
\hline
\multirow[c]{6}{*}{\rotatebox{90}{\parbox{2.85cm}{\textbf{MULTITuDE\_v3}}}} & Gemma-2-9b-it & \bfseries 0.9988 & \bfseries 0.9981 & 0.9883 & \bfseries 0.9997 & 0.9949 & \bfseries 0.9974 & \bfseries 0.9965 & 0.9895 & 0.9593 & 0.9942 & 0.9205 & 0.9962 & 0.9955 & 0.9906 & \bfseries 0.9995 & \bfseries 0.9980 & \bfseries 0.9979 & \bfseries 0.9945 & 0.9978 & 0.9991 & \bfseries 0.9963 & 0.9958 \\
& Qwen2-0.5B & 0.9578 & 0.9847 & 0.9919 & 0.9946 & 0.9888 & 0.9355 & 0.9800 & 0.9823 & 0.9529 & 0.9740 & 0.9407 & 0.9831 & 0.9752 & 0.9922 & 0.9887 & 0.9964 & 0.9932 & 0.9659 & 0.9921 & 0.9933 & 0.9754 & 0.9878 \\
& Qwen2-1.5B & 0.9930 & 0.9880 & 0.9972 & 0.9965 & 0.9871 & 0.9250 & 0.9937 & 0.9936 & 0.9633 & 0.9901 & 0.9661 & 0.9908 & 0.9850 & 0.9974 & 0.9941 & 0.9959 & 0.9943 & 0.9846 & 0.9975 & 0.9932 & 0.9815 & \bfseries 0.9991 \\
& Yi-1.5-6B & 0.9691 & 0.9666 & 0.9912 & 0.9905 & 0.9839 & 0.9194 & 0.9847 & 0.9804 & 0.9513 & 0.9833 & 0.9414 & 0.9823 & 0.9578 & 0.9929 & 0.9824 & 0.9835 & 0.9825 & 0.9591 & 0.9900 & 0.9890 & 0.9634 & 0.9939 \\
& mDeBERTa-v3-base & 0.9975 & 0.9976 & \bfseries 0.9981 & 0.9995 & \bfseries 0.9962 & 0.9919 & 0.9914 & \bfseries 0.9942 & \bfseries 0.9939 & 0.9917 & 0.9897 & 0.9983 & 0.9985 & \bfseries 0.9987 & 0.9972 & 0.9973 & 0.9970 & 0.9893 & \bfseries 0.9993 & 0.9994 & 0.9962 & 0.9983 \\
& mDeBERTa-v3-base (baseline) & 0.9961 & 0.9960 & 0.9960 & 0.9996 & 0.9904 & 0.9914 & 0.9803 & 0.9909 & 0.9927 & \bfseries 0.9966 & \bfseries 0.9919 & \bfseries 0.9990 & \bfseries 0.9986 & 0.9962 & 0.9969 & 0.9942 & 0.9962 & 0.9923 & 0.9929 & \bfseries 0.9995 & 0.9957 & 0.9907 \\
\hline
\multirow[c]{6}{*}{\rotatebox{90}{\parbox{2cm}{\textbf{MultiSocial}}}}& Gemma-2-9b-it & 0.9594 & 0.9853 & 0.9539 & 0.9774 & 0.9568 & 0.9357 & 0.9753 & 0.9694 & 0.9359 & 0.8861 & 0.8111 & 0.9755 & 0.9810 & 0.9575 & 0.9734 & 0.9623 & \bfseries 0.9743 & 0.9675 & 0.9650 & 0.9345 & 0.9519 & 0.9688 \\
& Qwen2-0.5B & 0.9549 & 0.9848 & 0.9634 & 0.9789 & 0.9527 & 0.9559 & 0.9717 & 0.9673 & 0.9744 & 0.8581 & 0.8223 & 0.9680 & 0.9836 & 0.9559 & 0.9604 & 0.9636 & 0.9499 & 0.9625 & 0.9604 & 0.8806 & 0.9343 & 0.9731 \\
& Qwen2-1.5B & 0.9673 & 0.9845 & 0.9652 & 0.9755 & 0.9581 & 0.9374 & 0.9767 & 0.9708 & 0.9230 & 0.8568 & 0.8284 & 0.9594 & 0.9555 & 0.9637 & 0.9645 & 0.9670 & 0.9426 & 0.9624 & 0.9599 & 0.8828 & 0.9395 & 0.9790 \\
& Yi-1.5-6B & 0.9342 & 0.9750 & 0.9562 & 0.9682 & 0.9446 & 0.9308 & 0.9674 & 0.9625 & 0.9658 & 0.8830 & 0.8422 & 0.9601 & 0.9707 & 0.9522 & 0.9579 & 0.9653 & 0.9238 & 0.9506 & 0.9233 & 0.8800 & 0.9287 & 0.9734 \\
& mDeBERTa-v3-base & \bfseries 0.9734 & 0.9887 & 0.9434 & 0.9635 & 0.9484 & 0.9611 & 0.9640 & 0.9543 & 0.9711 & 0.8378 & 0.7728 & 0.9572 & 0.9816 & 0.9507 & 0.9572 & 0.9505 & 0.9489 & 0.9563 & 0.9694 & \bfseries 0.9405 & 0.9561 & 0.9772 \\
& mDeBERTa-v3-base (baseline) & 0.9670 & \bfseries 0.9905 & \bfseries 0.9783 & \bfseries 0.9835 & \bfseries 0.9748 & \bfseries 0.9751 & \bfseries 0.9807 & \bfseries 0.9794 & \bfseries 0.9836 & \bfseries 0.9190 & \bfseries 0.8687 & \bfseries 0.9777 & \bfseries 0.9881 & \bfseries 0.9730 & \bfseries 0.9814 & \bfseries 0.9736 & 0.9736 & \bfseries 0.9695 & \bfseries 0.9711 & 0.9360 & \bfseries 0.9589 & \bfseries 0.9795 \\
\hline
\end{tabular}
}
\vspace{-2mm}
\caption{Per-language AUC ROC performance of the detectors on in-distribution data. The highest value in each dataset per each language is boldfaced.}
\label{tab:indistribution_lang}
\vspace{-3mm}
\end{table*}
\begin{table*}[!t]
\centering
\resizebox{\textwidth}{!}{
\addtolength{\tabcolsep}{-2pt}
\begin{tabular}{l|c||ccc|ccc||ccc|ccc||c}
\hline
 &  & \multicolumn{3}{c|}{\textbf{MultiSocial}} & \multicolumn{3}{c||}{\textbf{MULTITuDE\_v3}} & \multicolumn{3}{c|}{\textbf{MIX}} & \multicolumn{3}{c||}{\textbf{SemEval}} & \textbf{$\rightarrow$ Average} \\
\textbf{Detector} & \textbf{Threshold} & \textbf{FNR} & \textbf{FPR} & \textbf{Macro F1} & \textbf{FNR} & \textbf{FPR} & \textbf{Macro F1} & \textbf{FNR} & \textbf{FPR} & \textbf{Macro F1} & \textbf{FNR} & \textbf{FPR} & \textbf{Macro F1} & \textbf{Macro F1} \\
\hline
Gemma-2-9b-it & 0.500 & 0.157 & 0.052 & 0.767 & \textbf{0.026} & 0.038 & \textbf{0.942} & \textbf{0.141} & \textbf{0.222} & \textbf{0.818} & 0.128 & 0.088 & \textbf{0.891} & \textbf{0.855} \\
Qwen2-0.5B & 0.990 & 0.116 & 0.071 & 0.806 & 0.043 & 0.076 & 0.903 & 0.396 & 0.374 & 0.615 & 0.249 & 0.133 & 0.806 & 0.782 \\
Qwen2-1.5B & 0.900 & \textbf{0.101} & 0.109 & \textbf{0.813} & 0.068 & 0.039 & 0.877 & 0.299 & 0.353 & 0.674 & 0.187 & \textbf{0.038} & 0.884 & 0.812 \\
Yi-1.5-6B & 0.900 & 0.127 & 0.092 & 0.788 & 0.093 & 0.048 & 0.841 & 0.176 & 0.305 & 0.758 & 0.151 & 0.170 & 0.840 & 0.807 \\
mDeBERTa-v3-base & 0.970 & 0.130 & 0.043 & 0.798 & 0.033 & 0.022 & 0.935 & 0.489 & 0.296 & 0.604 & 0.256 & 0.133 & 0.803 & 0.785 \\
mDeBERTa-v3-base (baseline) & 0.999 & 0.169 & \textbf{0.030} & 0.761 & 0.036 & \textbf{0.018} & 0.932 & 0.520 & 0.402 & 0.537 & \textbf{0.076} & 0.375 & 0.774 & 0.751 \\
\hline
\end{tabular}
}
\vspace{-2mm}
\caption{Evaluation of detection performance using the calibrated threshold for each detector. The best achieved value of each metric is boldfaced (the lowest for FNR and FPR, the highest for Macro F1). The Gemma-2-9b-it detector is offering the best overall detection performance.}
\label{tab:calibrated}
\vspace{-5mm}
\end{table*}

As the results for out-of-distribution evaluation indicate (Table~\ref{tab:outofdistribution}), the baseline detector is the worst performing on both out-of-distribution datasets, which confirms our hypothesis about over-fitting. The better performance of robustly fine-tuned detectors using our augmented datasets for training clearly shows the benefits of our proposal.

The mDeBERTa-v3-base model (the same base model as the baseline) was able to achieve by 21\% higher AUC ROC performance on the MIX dataset and by 4\% on the SemEval dataset. Regarding the TPR @ 5\% FPR metric, the results show that the baseline detector was not able to achieve such a FPR for any classification threshold. On the other hand, the robustly fine-tuned version was able to detect almost 70\% of machine-generated texts on the SemEval dataset for 5\% FPR. It must be noted that not only the baseline, but also the Yi-1.5-6B and Qwen2-0.5B detectors have not detected any machine-generated text in the 5\% FPR conditions.
Overall, the best detector on out-of-distribution data is the Gemma-2-9b-it robustly fine-tuned model, detecting 52\% and 83\% of machine-generated texts of the MIX and SemEval datasets, respectively, for the 5\% FPR conditions.

\begin{figure}[!b]
\centering
\includegraphics[width=0.9\linewidth]{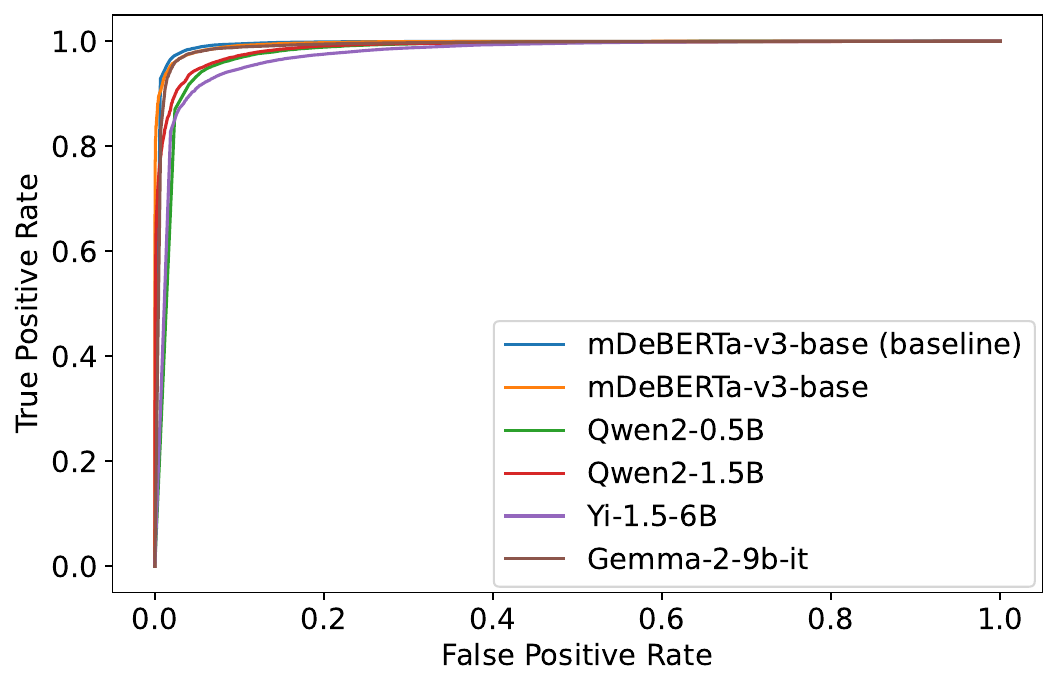}
\vspace{-3mm}
\caption{Receiver operating characteristic curves of the detectors using the MULTITuDE\_v3 dataset.}
\label{fig:roc_multitude3}
\vspace{-3mm}
\end{figure}

\begin{figure}[!t]
\centering
\includegraphics[width=0.9\linewidth]{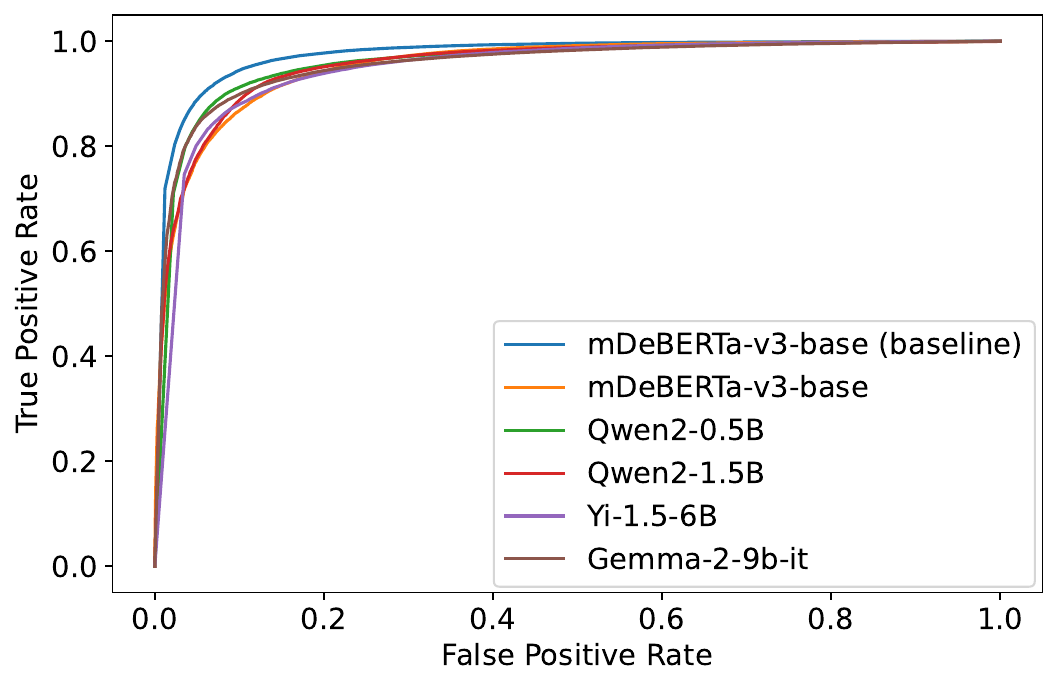}
\vspace{-3mm}
\caption{Receiver operating characteristic curves of the detectors using the MultiSocial dataset.}
\label{fig:roc_multisocial}
\vspace{-3mm}
\end{figure}

\begin{figure}[!t]
\centering
\includegraphics[width=0.9\linewidth]{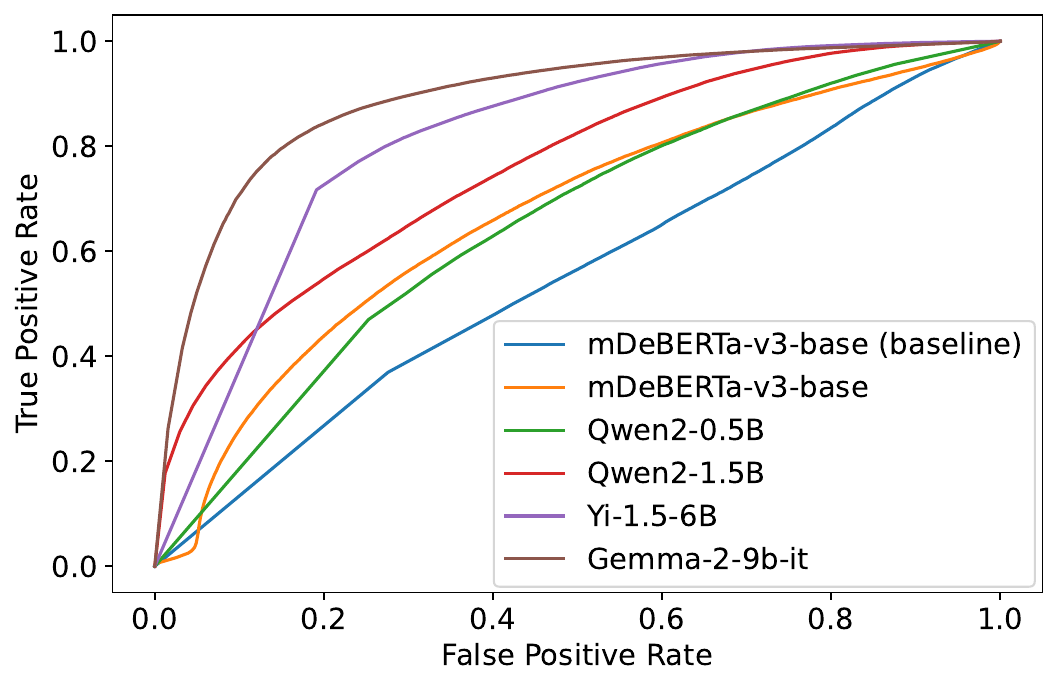}
\vspace{-3mm}
\caption{Receiver operating characteristic curves of the detectors using the MIX dataset.}
\label{fig:roc_mix}
\vspace{-3mm}
\end{figure}

\begin{figure}[!t]
\centering
\includegraphics[width=0.85\linewidth]{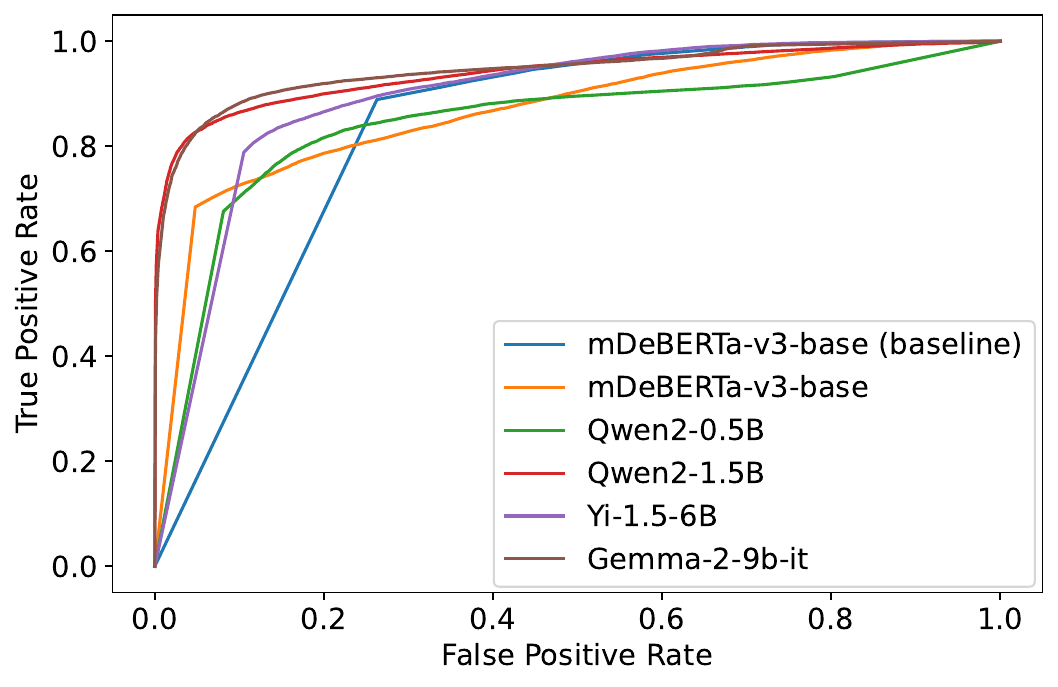}
\vspace{-3mm}
\caption{Receiver operating characteristic curves of the detectors using the SemEval dataset.}
\label{fig:roc_semeval}
\vspace{-3mm}
\end{figure}

A deeper analysis using the ROC curves of the detectors for individual datasets (\figurename~\ref{fig:roc_multitude3}--\ref{fig:roc_semeval}) revealed that the used out-of-distribution data are really challenging for them, while the in-distribution data are resulting in almost perfect classification capabilities. Figure~\ref{fig:roc_mix} shows that the performance of the baseline detector is only negligibly better than a random guessing (a diagonal) using the MIX dataset.


Based on such in-distribution and out-of-distribution evaluation, we have performed a manual classification-threshold calibration to achieve optimal trade-off detection (i.e., the highest Macro F1 across the datasets). The result for the calibrated thresholds are provided in Table~\ref{tab:calibrated}, where we provide the FNR, FPR, and Macro F1 metrics for each dataset, as well as the average of Macro F1 across the four datasets. The results indicate the significantly better performance of the Gemma-2-9b-it detector compared to the others, representing also the largest model. The Qwen2-1.5B detector offers also good performance, while being much smaller than the best model (9B vs 1.5B parameters $\sim 6 \times$ smaller); thus, offering a good trade-off between the performance and inference costs.

\section{Conclusions}

The fine-tuned machine-generated detection models are known to generalize poorly to out-of-distribution data. In this work, we have augmented training data to better balance the number of samples of machine and human classes. Besides, we have included obfuscated data directly into the detection model training; thus potentially making it more resistant against obfuscation as well. The results showed that such robustly fine-tuned detectors offer lower false positive rates and the proposed process is able to boost performance (AUC ROC) by up to 21\%. The performance of different detection models varies, providing an option to select the best suited one giving the trade-off between the detection performance and inference time (or availability of GPU accelerators).
The further work is required to optimize the hyper-parameters of the fine-tuning process to boost the detection performance of the best detection models even higher while offering even better generalizability to out-of-distribution data.

\newpage
\section*{Limitations}

Although we have tried to cover variety of domains, generators, and languages, the study is still limited in out-of-distribution evaluation. There is much more space to explore, and new text generation LLMs appear each month that could cause further performance degradation. We have also done a manual hyper-parameters optimization of fine-tuning process; however, we have covered just a small set of options, where a further tuning might increase the generalizability of the detectors.

\section*{Ethics Statement}
\label{sec:ethics}

Our work is focused on making the detection methods more robust to out-of-distribution data.
We use the existing datasets in our work in accordance with their intended use and licenses (for research purpose only). As a part of our work, we are not re-sharing any existing data or publishing any new dataset. For the research replicability and validation purposes, we are publishing the pre-processing, training, and evaluation source codes (for research purposes only). We have not used AI assistants in our work.

\section*{Acknowledgments}
This work was partially supported by the European Union under the Horizon Europe project AI-CODE, GA No. 101135437; by the EU NextGenerationEU through the Recovery and Resilience Plan for Slovakia under the project No. 09I01-03-V04-00059; and by the Slovak Research and Development Agency under the project Modermed, GA No. APVV-22-0414. We acknowledge EuroHPC Joint Undertaking for awarding us access to Leonardo at CINECA, Italy.

\bibliography{anthology,custom}

\begin{thebibliography}{36}
\providecommand{\natexlab}[1]{#1}

\bibitem[{Achiam et~al.(2023)Achiam, Adler, Agarwal, Ahmad, Akkaya, Aleman, Almeida, Altenschmidt, Altman, Anadkat et~al.}]{openai2023gpt4}
Josh Achiam, Steven Adler, Sandhini Agarwal, Lama Ahmad, Ilge Akkaya, Florencia~Leoni Aleman, Diogo Almeida, Janko Altenschmidt, Sam Altman, Shyamal Anadkat, et~al. 2023.
\newblock {GPT-4} technical report.
\newblock \emph{arXiv preprint arXiv:2303.08774}.

\bibitem[{Antoun et~al.(2023)Antoun, Mouilleron, Sagot, and Seddah}]{antoun-etal-2023-towards}
Wissam Antoun, Virginie Mouilleron, Beno{\^i}t Sagot, and Djam{\'e} Seddah. 2023.
\newblock \href {https://aclanthology.org/2023.jeptalnrecital-long.2/} {Towards a robust detection of language model-generated text: Is {C}hat{GPT} that easy to detect?}
\newblock In \emph{Actes de CORIA-TALN 2023. Actes de la 30e Conf{\'e}rence sur le Traitement Automatique des Langues Naturelles (TALN), volume 1 : travaux de recherche originaux -- articles longs}, pages 14--27, Paris, France. ATALA.

\bibitem[{Borji(2023)}]{borji2023categoricalarchivechatgptfailures}
Ali Borji. 2023.
\newblock \href {https://arxiv.org/abs/2302.03494} {A categorical archive of chatgpt failures}.
\newblock \emph{Preprint}, arXiv:2302.03494.

\bibitem[{Brown et~al.(2020)Brown, Mann, Ryder, Subbiah, Kaplan, Dhariwal, Neelakantan, Shyam, Sastry, Askell, Agarwal, Herbert-Voss, Krueger, Henighan, Child, Ramesh, Ziegler, Wu, Winter, Hesse, Chen, Sigler, Litwin, Gray, Chess, Clark, Berner, McCandlish, Radford, Sutskever, and Amodei}]{brown2020language}
Tom~B. Brown, Benjamin Mann, Nick Ryder, Melanie Subbiah, Jared Kaplan, Prafulla Dhariwal, Arvind Neelakantan, Pranav Shyam, Girish Sastry, Amanda Askell, Sandhini Agarwal, Ariel Herbert-Voss, Gretchen Krueger, Tom Henighan, Rewon Child, Aditya Ramesh, Daniel~M. Ziegler, Jeffrey Wu, Clemens Winter, Christopher Hesse, Mark Chen, Eric Sigler, Mateusz Litwin, Scott Gray, Benjamin Chess, Jack Clark, Christopher Berner, Sam McCandlish, Alec Radford, Ilya Sutskever, and Dario Amodei. 2020.
\newblock \href {https://arxiv.org/abs/2005.14165} {Language models are few-shot learners}.

\bibitem[{Chen et~al.(2022)Chen, Song, Wu, Wang, Xu, Chen, Zhou, and Li}]{chen-etal-2022-mtg}
Yiran Chen, Zhenqiao Song, Xianze Wu, Danqing Wang, Jingjing Xu, Jiaze Chen, Hao Zhou, and Lei Li. 2022.
\newblock \href {https://doi.org/10.18653/v1/2022.findings-naacl.192} {{MTG}: A benchmark suite for multilingual text generation}.
\newblock In \emph{Findings of the Association for Computational Linguistics: NAACL 2022}, pages 2508--2527, Seattle, United States. Association for Computational Linguistics.

\bibitem[{Crothers et~al.(2023)Crothers, Japkowicz, and Viktor}]{crothers2023machine}
Evan~N Crothers, Nathalie Japkowicz, and Herna~L Viktor. 2023.
\newblock Machine-generated text: A comprehensive survey of threat models and detection methods.
\newblock \emph{IEEE Access}, 11:70977--71002.

\bibitem[{Dettmers et~al.(2024)Dettmers, Pagnoni, Holtzman, and Zettlemoyer}]{dettmers2024qlora}
Tim Dettmers, Artidoro Pagnoni, Ari Holtzman, and Luke Zettlemoyer. 2024.
\newblock {QLoRA}: Efficient finetuning of quantized {LLM}s.
\newblock \emph{Advances in Neural Information Processing Systems}, 36.

\bibitem[{Fagni et~al.(2021)Fagni, Falchi, Gambini, Martella, and Tesconi}]{fagni2021tweepfake}
Tiziano Fagni, Fabrizio Falchi, Margherita Gambini, Antonio Martella, and Maurizio Tesconi. 2021.
\newblock {TweepFake}: About detecting deepfake tweets.
\newblock \emph{Plos one}, 16(5):e0251415.

\bibitem[{Fan et~al.(2018)Fan, Lewis, and Dauphin}]{fan-etal-2018-hierarchical}
Angela Fan, Mike Lewis, and Yann Dauphin. 2018.
\newblock \href {https://doi.org/10.18653/v1/P18-1082} {Hierarchical neural story generation}.
\newblock In \emph{Proceedings of the 56th Annual Meeting of the Association for Computational Linguistics (Volume 1: Long Papers)}, pages 889--898, Melbourne, Australia. Association for Computational Linguistics.

\bibitem[{Gabriel et~al.(2024)Gabriel, Lyu, Siderius, Ghassemi, Andreas, and Ozdaglar}]{gabriel2024generative}
Saadia Gabriel, Liang Lyu, James Siderius, Marzyeh Ghassemi, Jacob Andreas, and Asu Ozdaglar. 2024.
\newblock Generative ai in the era of'alternative facts'.

\bibitem[{He et~al.(2024)He, Shen, Chen, Backes, and Zhang}]{he2024mgtbench}
Xinlei He, Xinyue Shen, Zeyuan Chen, Michael Backes, and Yang Zhang. 2024.
\newblock {MGTBench}: Benchmarking machine-generated text detection.
\newblock In \emph{Proceedings of the 2024 on ACM SIGSAC Conference on Computer and Communications Security}, pages 2251--2265.

\bibitem[{Kaddour et~al.(2023)Kaddour, Harris, Mozes, Bradley, Raileanu, and McHardy}]{kaddour2023challenges}
Jean Kaddour, Joshua Harris, Maximilian Mozes, Herbie Bradley, Roberta Raileanu, and Robert McHardy. 2023.
\newblock \href {https://arxiv.org/abs/2307.10169} {Challenges and applications of large language models}.
\newblock \emph{arXiv preprint arXiv:2307.10169}.

\bibitem[{Lee et~al.(2022)Lee, Liang, and Yang}]{10.1145/3491102.3502030}
Mina Lee, Percy Liang, and Qian Yang. 2022.
\newblock \href {https://doi.org/10.1145/3491102.3502030} {Coauthor: Designing a human-ai collaborative writing dataset for exploring language model capabilities}.
\newblock In \emph{Proceedings of the 2022 CHI Conference on Human Factors in Computing Systems}, CHI '22, New York, NY, USA. Association for Computing Machinery.

\bibitem[{Li et~al.(2024)Li, Li, Cui, Bi, Wang, Wang, Yang, Shi, and Zhang}]{li-etal-2024-mage}
Yafu Li, Qintong Li, Leyang Cui, Wei Bi, Zhilin Wang, Longyue Wang, Linyi Yang, Shuming Shi, and Yue Zhang. 2024.
\newblock \href {https://doi.org/10.18653/v1/2024.acl-long.3} {{MAGE}: Machine-generated text detection in the wild}.
\newblock In \emph{Proceedings of the 62nd Annual Meeting of the Association for Computational Linguistics (Volume 1: Long Papers)}, pages 36--53, Bangkok, Thailand. Association for Computational Linguistics.

\bibitem[{Lucas et~al.(2023)Lucas, Uchendu, Yamashita, Lee, Rohatgi, and Lee}]{lucas-etal-2023-fighting}
Jason Lucas, Adaku Uchendu, Michiharu Yamashita, Jooyoung Lee, Shaurya Rohatgi, and Dongwon Lee. 2023.
\newblock \href {https://doi.org/10.18653/v1/2023.emnlp-main.883} {Fighting fire with fire: The dual role of {LLM}s in crafting and detecting elusive disinformation}.
\newblock In \emph{Proceedings of the 2023 Conference on Empirical Methods in Natural Language Processing}, pages 14279--14305, Singapore. Association for Computational Linguistics.

\bibitem[{Macko et~al.(2024{\natexlab{a}})Macko, Kopal, Moro, and Srba}]{multisocial}
Dominik Macko, Jakub Kopal, Robert Moro, and Ivan Srba. 2024{\natexlab{a}}.
\newblock \href {https://arxiv.org/abs/2406.12549} {Multi{S}ocial: Multilingual benchmark of machine-generated text detection of social-media texts}.
\newblock \emph{Preprint}, arXiv:2406.12549.

\bibitem[{Macko et~al.(2023)Macko, Moro, Uchendu, Lucas, Yamashita, Pikuliak, Srba, Le, Lee, Simko, and Bielikova}]{macko-etal-2023-multitude}
Dominik Macko, Robert Moro, Adaku Uchendu, Jason Lucas, Michiharu Yamashita, Mat{\'u}{\v{s}} Pikuliak, Ivan Srba, Thai Le, Dongwon Lee, Jakub Simko, and Maria Bielikova. 2023.
\newblock \href {https://doi.org/10.18653/v1/2023.emnlp-main.616} {{MULTIT}u{DE}: Large-scale multilingual machine-generated text detection benchmark}.
\newblock In \emph{Proceedings of the 2023 Conference on Empirical Methods in Natural Language Processing}, pages 9960--9987, Singapore. Association for Computational Linguistics.

\bibitem[{Macko et~al.(2024{\natexlab{b}})Macko, Moro, Uchendu, Srba, Lucas, Yamashita, Tripto, Lee, Simko, and Bielikova}]{macko-etal-2024-authorship}
Dominik Macko, Robert Moro, Adaku Uchendu, Ivan Srba, Jason~S Lucas, Michiharu Yamashita, Nafis~Irtiza Tripto, Dongwon Lee, Jakub Simko, and Maria Bielikova. 2024{\natexlab{b}}.
\newblock \href {https://doi.org/10.18653/v1/2024.findings-emnlp.369} {Authorship obfuscation in multilingual machine-generated text detection}.
\newblock In \emph{Findings of the Association for Computational Linguistics: EMNLP 2024}, pages 6348--6368, Miami, Florida, USA. Association for Computational Linguistics.

\bibitem[{Merity et~al.(2016)Merity, Xiong, Bradbury, and Socher}]{merity2016pointer}
Stephen Merity, Caiming Xiong, James Bradbury, and Richard Socher. 2016.
\newblock \href {https://arxiv.org/abs/1609.07843} {Pointer sentinel mixture models}.
\newblock \emph{Preprint}, arXiv:1609.07843.

\bibitem[{Pu et~al.(2023)Pu, Sarwar, Abdullah, Rehman, Kim, Bhattacharya, Javed, and Viswanath}]{pu2023deepfake}
Jiameng Pu, Zain Sarwar, Sifat~Muhammad Abdullah, Abdullah Rehman, Yoonjin Kim, Parantapa Bhattacharya, Mobin Javed, and Bimal Viswanath. 2023.
\newblock Deepfake text detection: Limitations and opportunities.
\newblock In \emph{2023 IEEE symposium on security and privacy (SP)}, pages 1613--1630. IEEE.

\bibitem[{Pu et~al.(2022)Pu, Huang, Xi, Chen, Chen, and Zhang}]{pu-etal-2022-unraveling}
Jiashu Pu, Ziyi Huang, Yadong Xi, Guandan Chen, Weijie Chen, and Rongsheng Zhang. 2022.
\newblock \href {https://aclanthology.org/2022.lrec-1.744/} {Unraveling the mystery of artifacts in machine generated text}.
\newblock In \emph{Proceedings of the Thirteenth Language Resources and Evaluation Conference}, pages 6889--6898, Marseille, France. European Language Resources Association.

\bibitem[{Radford et~al.(2019)Radford, Wu, Child, Luan, Amodei, Sutskever et~al.}]{radford2019language}
Alec Radford, Jeffrey Wu, Rewon Child, David Luan, Dario Amodei, Ilya Sutskever, et~al. 2019.
\newblock Language models are unsupervised multitask learners.
\newblock \emph{OpenAI blog}, 1(8):9.

\bibitem[{Rosati(2022)}]{rosati-2022-synscipass}
Domenic Rosati. 2022.
\newblock \href {https://aclanthology.org/2022.sdp-1.27/} {{S}yn{S}ci{P}ass: detecting appropriate uses of scientific text generation}.
\newblock In \emph{Proceedings of the Third Workshop on Scholarly Document Processing}, pages 214--222, Gyeongju, Republic of Korea. Association for Computational Linguistics.

\bibitem[{Sadasivan et~al.(2025)Sadasivan, Kumar, Balasubramanian, Wang, and Feizi}]{sadasivan2025can}
Vinu~Sankar Sadasivan, Aounon Kumar, Sriram Balasubramanian, Wenxiao Wang, and Soheil Feizi. 2025.
\newblock \href {https://openreview.net/forum?id=OOgsAZdFOt} {Can {AI}-generated text be reliably detected? stress testing {AI} text detectors under various attacks}.
\newblock \emph{Transactions on Machine Learning Research}.

\bibitem[{Sarvazyan et~al.(2023)Sarvazyan, Gonz{\'a}lez, Franco-Salvador, Rangel, Chulvi, and Rosso}]{sarvazyan2023overview}
Areg~Mikael Sarvazyan, Jos{\'e}~{\'A}ngel Gonz{\'a}lez, Marc Franco-Salvador, Francisco Rangel, Berta Chulvi, and Paolo Rosso. 2023.
\newblock Overview of {AuTexTification} at {IberLEF} 2023: Detection and attribution of machine-generated text in multiple domains.
\newblock \emph{arXiv preprint arXiv:2309.11285}.

\bibitem[{See et~al.(2017)See, Liu, and Manning}]{see-etal-2017-get}
Abigail See, Peter~J. Liu, and Christopher~D. Manning. 2017.
\newblock \href {https://doi.org/10.18653/v1/P17-1099} {Get to the point: Summarization with pointer-generator networks}.
\newblock In \emph{Proceedings of the 55th Annual Meeting of the Association for Computational Linguistics (Volume 1: Long Papers)}, pages 1073--1083, Vancouver, Canada. Association for Computational Linguistics.

\bibitem[{Shamardina et~al.(2022)Shamardina, Mikhailov, Chernianskii, Fenogenova, Saidov, Valeeva, Shavrina, Smurov, Tutubalina, and Artemova}]{Shamardina_2022}
Tatiana Shamardina, Vladislav Mikhailov, Daniil Chernianskii, Alena Fenogenova, Marat Saidov, Anastasiya Valeeva, Tatiana Shavrina, Ivan Smurov, Elena Tutubalina, and Ekaterina Artemova. 2022.
\newblock \href {https://doi.org/10.28995/2075-7182-2022-21-497-511} {Findings of the the {RuATD} shared task 2022 on artificial text detection in russian}.
\newblock In \emph{Computational Linguistics and Intellectual Technologies}, page 497–511. RSUH.

\bibitem[{Tan et~al.(2020)Tan, Plummer, and Saenko}]{tan-etal-2020-detecting}
Reuben Tan, Bryan Plummer, and Kate Saenko. 2020.
\newblock \href {https://doi.org/10.18653/v1/2020.emnlp-main.163} {Detecting cross-modal inconsistency to defend against neural fake news}.
\newblock In \emph{Proceedings of the 2020 Conference on Empirical Methods in Natural Language Processing (EMNLP)}, pages 2081--2106, Online. Association for Computational Linguistics.

\bibitem[{Uchendu et~al.(2021)Uchendu, Ma, Le, Zhang, and Lee}]{uchendu-etal-2021-turingbench-benchmark}
Adaku Uchendu, Zeyu Ma, Thai Le, Rui Zhang, and Dongwon Lee. 2021.
\newblock \href {https://doi.org/10.18653/v1/2021.findings-emnlp.172} {{TURINGBENCH}: A benchmark environment for {T}uring test in the age of neural text generation}.
\newblock In \emph{Findings of the Association for Computational Linguistics: EMNLP 2021}, pages 2001--2016, Punta Cana, Dominican Republic. Association for Computational Linguistics.

\bibitem[{Verma et~al.(2021)Verma, Agrawal, Amorim, and Prodan}]{9395133}
Pawan~Kumar Verma, Prateek Agrawal, Ivone Amorim, and Radu Prodan. 2021.
\newblock \href {https://doi.org/10.1109/TCSS.2021.3068519} {{WELFake}: Word embedding over linguistic features for fake news detection}.
\newblock \emph{IEEE Transactions on Computational Social Systems}, 8(4):881--893.

\bibitem[{Vykopal et~al.(2024)Vykopal, Pikuliak, Srba, Moro, Macko, and Bielikova}]{vykopal-etal-2024-disinformation}
Ivan Vykopal, Mat{\'u}{\v{s}} Pikuliak, Ivan Srba, Robert Moro, Dominik Macko, and Maria Bielikova. 2024.
\newblock \href {https://doi.org/10.18653/v1/2024.acl-long.793} {Disinformation capabilities of large language models}.
\newblock In \emph{Proceedings of the 62nd Annual Meeting of the Association for Computational Linguistics (Volume 1: Long Papers)}, pages 14830--14847, Bangkok, Thailand. Association for Computational Linguistics.

\bibitem[{Wahle et~al.(2022)Wahle, Ruas, Kirstein, and Gipp}]{wahle-etal-2022-large}
Jan~Philip Wahle, Terry Ruas, Frederic Kirstein, and Bela Gipp. 2022.
\newblock \href {https://doi.org/10.18653/v1/2022.emnlp-main.62} {How large language models are transforming machine-paraphrase plagiarism}.
\newblock In \emph{Proceedings of the 2022 Conference on Empirical Methods in Natural Language Processing}, pages 952--963, Abu Dhabi, United Arab Emirates. Association for Computational Linguistics.

\bibitem[{Wang et~al.(2024)Wang, Mansurov, Ivanov, Su, Shelmanov, Tsvigun, Mohammed~Afzal, Mahmoud, Puccetti, and Arnold}]{wang-etal-2024-semeval-2024}
Yuxia Wang, Jonibek Mansurov, Petar Ivanov, Jinyan Su, Artem Shelmanov, Akim Tsvigun, Osama Mohammed~Afzal, Tarek Mahmoud, Giovanni Puccetti, and Thomas Arnold. 2024.
\newblock \href {https://doi.org/10.18653/v1/2024.semeval-1.279} {{S}em{E}val-2024 task 8: Multidomain, multimodel and multilingual machine-generated text detection}.
\newblock In \emph{Proceedings of the 18th International Workshop on Semantic Evaluation (SemEval-2024)}, pages 2057--2079, Mexico City, Mexico. Association for Computational Linguistics.

\bibitem[{Zellers et~al.(2019)Zellers, Holtzman, Rashkin, Bisk, Farhadi, Roesner, and Choi}]{zellers2019defending}
Rowan Zellers, Ari Holtzman, Hannah Rashkin, Yonatan Bisk, Ali Farhadi, Franziska Roesner, and Yejin Choi. 2019.
\newblock Defending against neural fake news.
\newblock \emph{Advances in neural information processing systems}, 32.

\bibitem[{Zhuo et~al.(2023)Zhuo, Huang, Chen, and Xing}]{zhuo2023redteamingchatgptjailbreaking}
Terry~Yue Zhuo, Yujin Huang, Chunyang Chen, and Zhenchang Xing. 2023.
\newblock \href {https://arxiv.org/abs/2301.12867} {Red teaming {ChatGPT} via jailbreaking: Bias, robustness, reliability and toxicity}.
\newblock \emph{Preprint}, arXiv:2301.12867.

\bibitem[{Zugecova et~al.(2024)Zugecova, Macko, Srba, Moro, Kopal, Marcincinova, and Mesarcik}]{zugecova2024evaluationllmvulnerabilitiesmisused}
Aneta Zugecova, Dominik Macko, Ivan Srba, Robert Moro, Jakub Kopal, Katarina Marcincinova, and Matus Mesarcik. 2024.
\newblock \href {https://arxiv.org/abs/2412.13666} {Evaluation of {LLM} vulnerabilities to being misused for personalized disinformation generation}.
\newblock \emph{Preprint}, arXiv:2412.13666.

\end{thebibliography}

\appendix

\section{Computational Resources}
\label{sec:compute}

For experiments regarding model fine-tuning and inference processes, we have used 1x A100 64GB GPU, cumulatively consuming around 1,000 GPU-core hours. For data augmentation, pre-processing, combining, and analysis of the results, we have used Jupyter Lab running on 4 CPU cores, without the GPU acceleration.

\section{MIX Dataset}
\label{sec:appendix_mix}

For the diverse mixture dataset of existing labeled datasets for out-of-distribution evaluation, we have combined the samples of data (up to 5,000 text samples per language/source and label) of the following datasets:
TuringBench \citep{uchendu-etal-2021-turingbench-benchmark}, Grover \citep{zellers2019defending}, GPT-2 \citep{radford2019language}, GPT-3 \citep{brown2020language}, Paired \citep{pu-etal-2022-unraveling}, CNN/DailyMail \citep{see-etal-2017-get}, NeuralNews \citep{tan-etal-2020-detecting}, InTheWild \citep{pu2023deepfake}, Wikitext \citep{merity2016pointer}, WritingPrompts \citep{fan-etal-2018-hierarchical}, CoAuthor \citep{10.1145/3491102.3502030}, TweepFake \citep{fagni2021tweepfake}, WELFake \citep{9395133}, AuTexTification \citep{sarvazyan2023overview}, MTG \citep{chen-etal-2022-mtg}, MGTBench \citep{he2024mgtbench}, RuATD \citep{Shamardina_2022}, SynSciPass \citep{rosati-2022-synscipass}.

The data have been combined in a common way, providing texts, labels, source, and if available also the language and domain/topic information. Afterwards, the duplicates of the texts have been removed and the final dataset has been pseudo-randomly sampled to include 100k samples for the human and machine classes. For the MIX2k validation dataset, the 1k samples of each class have been pseudo-randomly sampled.

Due to various licensing of the data of the used datasets (some of which are also composites of multiple data sources), we are not re-sharing the dataset itself. However, for the research purpose (validation and replication), we are providing the data construction source code.

\end{document}